\documentclass[conference]{IEEEtran}

\usepackage{geometry}
\usepackage{amsmath,graphicx} %spconf, 
\usepackage{dsfont}
\usepackage{amssymb}
\usepackage{amsfonts}
\usepackage[all,poly]{xy}
\usepackage{multirow}
\usepackage{color}
\usepackage[noadjust]{cite}
\usepackage{subfigure}
\usepackage{mathrsfs}
\usepackage{algorithmic}
\usepackage[linesnumbered,lined,ruled]{algorithm2e}
\usepackage{url}
\usepackage{lipsum}
\usepackage{stfloats}
\usepackage{epsfig}
\usepackage{amsthm}
\usepackage{gensymb}

\usepackage{epstopdf}

\newcommand{\bs}{\boldsymbol}

\newcommand{\tcpp}[1]{\tcp{\emph{\small{#1}}}}

\newcommand{\minimize}[2]{\ensuremath{\underset{\substack{{#1}}}%
{\text{\rm minimize}}\;\;#2 }}
\newcommand{\prox}{\ensuremath{\text{\rm prox}}}

\def\bfh{{\mathbf{h}}}

\def\bfm{{\mathbf{m}}}

\def\bfw{{\mathbf{w}}}

\def\bfD{{\mathbf{D}}}

\def\bfF{{\mathbf{F}}}
\def\bfG{{\mathbf{G}}}
\def\bfH{{\mathbf{H}}}
\def\bfI{{\mathbf{I}}}

\def\bfM{{\mathbf{M}}}
\def\bfN{{\mathbf{N}}}

\def\bfQ{{\mathbf{Q}}}

\def\bfU{{\mathbf{U}}}

\def\bfW{{\mathbf{W}}}

\def\bfY{{\mathbf{Y}}}

\def\calC{{\mathcal{C}}}

\def\calN{{\mathcal{N}}}
\def\calO{{\mathcal{O}}}

\def\bsm{{\boldsymbol{m}}}

\newcounter{algo}
\renewcommand{\thealgo}{\arabic{algo}}

\graphicspath{{figures/}}

\def\ninept{\def\baselinestretch{.95}\let\normalsize\small\normalsize}

\begin{document}
%\ninept

%\title{Least Squares Regression with Gaussian Markov Random Field Regularization}
\title{A Fast Algorithm Based on a Sylvester-like Equation for LS Regression with GMRF Prior}

\author{\IEEEauthorblockN{Qi Wei}
\IEEEauthorblockA{
Dept. of ECE \\ %Electr. and Comput. Eng.
Duke University\\
Durham, USA 27710\\
qi.wei@duke.edu \\}
\and 
\IEEEauthorblockN{Emilie Chouzenoux}
\IEEEauthorblockA{
LIGM, Universit\'{e} \\
Paris-Est Marne-la-Vall\'{e}e\\
Champs-sur-Marne, France 77454\\
emilie.chouzenoux@univ-mlv.fr}
\and
\IEEEauthorblockN{Jean-Yves Tourneret}
\IEEEauthorblockA{
IRIT/INP-ENSEEIHT\\
Universit\'e de Toulouse\\
Toulouse, France 31071\\
jyt@enseeiht.fr \\
}
\and 
\IEEEauthorblockN{Jean-Christophe Pesquet}
\IEEEauthorblockA{
Center for Visual Computing\\
CentraleSupelec-University Paris-Saclay\\ 
France 92295\\
jean-christophe@pesquet.eu}
}

%\author{\IEEEauthorblockN{Qi Wei, \IEEEmembership{Member,~IEEE},
%Emilie Chouzenoux, \IEEEmembership{Member,~IEEE}, %\\ % single column
%Jean-Yves Tourneret, \IEEEmembership{Senior Member,~IEEE}, 
%and Jean-Christophe Pesquet, \IEEEmembership{Fellow,~IEEE}}
%\thanks{Qi Wei is with Department of Electrical and Computer Engineering, Duke University, Durham, USA 27710
%(e-mail: {qi.wei@duke.edu}).}
%\thanks{Emilie Chouzenoux is with LIGM, Universit\'{e} Paris-Est Marne-la-Vall\'{e}e, Champs-sur-Marne, France 77454
%(e-mail: {emilie.chouzenoux@univ-mlv.fr}).}
%\thanks{Jean-Yves Tourneret is with University of Toulouse, IRIT/INP-ENSEEIHT, 2 rue Camichel, BP 7122, Toulouse cedex 7,
%France 31071 (e-mail: {jean-yves.tourneret@enseeiht.fr}).}
%\thanks{Jean-Christophe Pesquet is with Center for Visual Computing, CentraleSupelec-University Paris-Saclay, France 92295
%(e-mail: {jean-christophe@pesquet.eu}).}}

\maketitle

\begin{abstract}
This paper presents a fast approach for penalized least squares (LS) regression problems using 
a 2D Gaussian Markov random field (GMRF) prior.
% based on a closed-form  solution of a matrix equation. 
More precisely,  the computation of the proximity operator of the LS criterion regularized by
different GMRF potentials is formulated as solving a Sylvester-like matrix equation. By exploiting the structural properties
of GMRFs, this matrix equation is solved column-wise in an analytical way.
The proposed algorithm can be embedded into a wide range of proximal algorithms to
solve LS regression problems including a convex penalty.
Experiments carried out in the case of a constrained LS regression problem arising in 
a multichannel image processing application, provide evidence that an alternating direction 
method of multipliers performs quite efficiently in this context.
\end{abstract}

\IEEEpeerreviewmaketitle

%\begin{keywords}
%least squares regression, Gaussian Markov random field, Sylvester equation, closed-form solution, proximity operator
%\end{keywords}

\section{Introduction}
Constrained or penalized least squares (LS) problems have been widely encountered in various 
signal/image processing applications, such as spectral unmixing \cite{Keshava2002, Chouzenoux2014,Wei2015FastUnmixing},
supervised source separation \cite{Dobigeon2009}, image classification \cite{Chang1998b}, material 
quantification \cite{Wang2006} or subpixel detection \cite{Manolakis2001}. The LS problem 
often results from the following linear model which has been successfully  used in the applications
mentioned above:
\begin{equation}
\bf Y= WH+N
\label{eq:linear_model}
\end{equation}
where $\bfY \in \mathbb{R}^{m \times n}$ is the observed data matrix (each row of $\bfY$ is the vectorized version of an image),
$\bfW \in \mathbb{R}^{m \times d}$ is a basis matrix that will be assumed to be known in this work,
$\bfH \in \mathbb{R}^{d \times n}$ is a matrix containing the regression coefficients,
and $\bfN \in \mathbb{R}^{m \times n}$ is the noise term which can be assumed to
follow a multivariate Gaussian distribution.
Note that LS can be classically interpreted as projecting the observed data
onto the subspace spanned by the columns of $\bfW$.

As the LS problem associated with \eqref{eq:linear_model} is usually ill-posed, 
e.g., some columns of $\bfW$ may be similar, it is necessary to introduce 
priors/regularizations for $\bfH$ to make the problem well-conditioned~\cite{Lawson1974}. %Aside from the data term
Enforcing spatial regularization on the matrix $\bfH$ is a strategy for incorporating prior information,
e.g., total variation (TV), Markov random field (MRF) penalty, sparsity constraints in the wavelet domain, etc.
Among these, a powerful and important way of exploiting the correlations between pixels of an image
is to consider Gaussian Markov random fields (GMRFs), which have been extensively used
in image processing applications such as denoising \cite{Malfait1997}, super-resolution \cite{Kasetkasem2005}, 
segmentation \cite{Elia2003} and spectral unmixing \cite{Eches2013adaptive}.
%which are discrete domain Gaussian random fields equipped with a Markov property 
%and have been widely used in different areas of spatial statistics mentioned as above.
%As is well known, Gaussian Markov random fields (GMRFs) is a powerful and important tools for modeling 
%spatial data/image, which .
Constructing a GMRF amounts to define a finite-dimensional random vector with
a multivariate normal distribution having nontrivial conditional Markov dependence properties. 
GMRFs allow us to exploit analytical results obtained for the Gaussian distribution
and to enforce Markovian properties, leading to computationally efficient algorithms.
%as well as Markov properties, thus allow for the development of computationally efficient algorithms.
In general, different images can be characterized by GMRF distributions with different parameters. 
For example, the distributions of water and soil in a remote sensing image can be 
modeled by two different GMRF distributions based upon their physical locations.
Mathematically, the GMRF regularizations associated with the two rows of $\bfH$
corresponding to water and soil should obviously be different. This diversity makes the
corresponding optimization problem quite challenging, leading to the solution of a tensor equation. 
A number of efficient sampling algorithms such as those based on Markov chain Monte 
Carlo (MCMC) algorithms have been designed for statistical inference, which are
effective but generally time consuming \cite{Rue2005GMRF, Eches2013adaptive}.

In this paper, we adopt a proximal approach \cite{Combettes2011} to address this variational problem.
We start by showing that the computation of the proximity operator of the 
LS criterion with GMRF regularization can be performed by
solving a Sylvester-like matrix equation and propose an algorithm to solve it analytically
by taking advantage of the properties of stationary 2D GMRFs. More specifically, the block circulant 
properties of the covariance matrix of such a field is exploited to simplify the associated
matrix equation. The resulting closed-form solution is easy to implement and very fast to compute.

This paper is organized as follows. Section \ref{sec:prob_form} formulates the
regularized LS regression for the considered class of linear models and GMRF priors.
Section \ref{se:prox} addresses the problem of computing the associated proximity operator
by solving in a fast manner a Sylvester-like matrix equation.
Section \ref{se:penLS} shows the benefit of this approach for solving more challenging convex optimization problems. 
Simulation results are presented in Section \ref{sec:simu} showing the good performance of the proposed approach,
whereas conclusions are reported in Section \ref{sec:conclusion}.
% It is convenient and invaluable to combine 
%the analytical results for the Gaussian distribution and the Markov properties, which enables us to solve 
%a large class of statistical models. 

%On the other hand, the Markov property has become a requirement for constructing efficient Markov chain Monte Carlo (MCMC) algorithms for GMRFs. Rue [3] showed that the Markov property makes it possible to apply numerical methods on sparse matrices. He proposed fast algorithms for sampling and evaluating the log-density of a GMRF, and conducted efficient MCMC-based inferences. Rue and Held [4] provides a comprehensive account of the main properties of GMRFs, emphasizes the strong connection between GMRFs and numerical methods for sparse matrices, and outlines various applications of GMRFs for statistical inference (e.g., spatial statistics, time-series analysis, graphical models).

\section{Problem formulation}
\label{sec:prob_form}
\subsection{Observation model}
Decomposing the matrices $\bfW$ and $\bfH$ as $\bfW = [\bfw_1, \cdots, \bfw_d]$ and $\bfH^T = [\bfh_1, \cdots, \bfh_d]$,
where $\bfw_k$ is the $k$th column of $\bfW$ and $\bfh_i^T$ is the $i$th row of $\bfH$,
\eqref{eq:linear_model} can be rewritten as 
\begin{equation}
\bfY=  \sum_{i=1}^{d} \bfw_i \bfh_i^T+ \bfN.
\end{equation}
Note that each pixel (column) of the image (matrix) $\bfY$ is
the linear combination of $d$ basis vectors $\bfw_1, \cdots, \bfw_d$ (e.g., 
$d$ materials whose signatures are the columns of 
$\bfW$). Estimating the matrix $\bfH$ from the observed matrix $\bfY$
with possible constraints about the vectors $\bfh_i$ is a classical LS
problem that has been considered in particular in source separation \cite{Zibulevsky2001} and spectral unmixing \cite{Keshava2002, Heinz2001}.
%The LS problem is challenging when the basis matrix $\bfW$ is ill-conditioned, which will be addressed in this paper.
% This problem has been encountered in source separation, spectral unmixing, wireless communication, etc.
\subsection{Gaussian Markov Random Fields}
According to the Hammersley-Clifford theorem \cite{Clifford1990, Rue2005GMRF}, an MRF can equivalently 
be characterized by a Gibbs distribution. More specifically, a zero-mean Gaussian random field 
%$\bfh \sim \calN (\bs{\mu}, \bs{\Sigma})$
$(h_k)_{1 \leq k \leq n}  \in \mathbb{R}$
satisfying\footnote{To simplify notation, the index of $\bfh_i$ has been dropped in this section.}
\begin{equation}
p(h_k \mid {h_\ell, \ell \neq k})=p(h_k \mid {h_\ell, \ell \in \calN_k})
\end{equation}
where $\calN_k$ contains the neighbors of the $k$th element $h_k$, is 
a GMRF. The distribution of $\bfh = [h_1,\cdots,h_k]^T$ can be written as
\begin{equation}
p(\bfh)=\frac{1}{c}\exp{\left(-\frac{\lambda }{2}\sum_{k=1}^{n}\big(h_k-\sum_{\ell\in \calN_k} \alpha_\ell h_\ell\big)^2 \right)}
\label{e:GMRFdetailed}
\end{equation} 
where $\lambda>0$ is a scale parameter and the normalizing constant $c>0$ is the partition function of this probability distribution, which is generally unknown. 
Equivalently, \eqref{e:GMRFdetailed} reads
\begin{equation}
p(\bfh)=\frac{1}{c}\exp{\left(-\frac{\lambda}{2}\|\bfh-\bfQ\bfh\|_2^2 \right)}
\label{eq:GMRF}
\end{equation}
where $\lambda(\bfI-\bfQ)^T(\bfI-\bfQ)$ is the precision matrix, $\bfI$ denotes the identity matrix and, in the 2D stationary case with periodic boundary condition, $\bfQ$ is a block circulant matrix with circulant blocks (BCCB) with its first column 
%defined as the Fourier transform of 
built from the coefficient vector $\bs{\alpha}=\left(\alpha_1,\ldots,\alpha_q\right)^T$,
$q = | \calN_k | $ being the number of elements in the neighborhood of $h_k$. 

%\begin{equation}
%\phi(\bfH) =\sum_{i=1}^d \frac{1}{2}\|\bfh_i-\bfh_i\bfQ_i\|_F^2
%\end{equation}

\section{Fast computation of the proximity operator of the least squares criterion with GMRF prior}
\label{se:prox}
Assuming that the columns of $\bfH$ are independent and assigned a GMRF prior and considering the 
likelihood term from \eqref{eq:linear_model}
%Incorporating the likelihood term from \eqref{eq:linear_model} and the GMRF prior from \eqref{eq:GMRF} 
%for each of the rows of $\bfH$, here assumed to be independent,
leads to the following LS regression problem: 
\begin{equation}
\label{eq:prob_LS}
\minimize{\bfH \in \mathbb{R}^{d\times n}} f(\bfH)
\end{equation}
where 
\begin{equation*}
f(\bfH) = \frac{1}{2}\|\bfY-\sum_{i=1}^{d} \bfw_i \bfh_i^T\|_{\rm F}^2 +\sum_{i=1}^d \frac{\lambda_i}{2}\|\bfh_i^T - \bfh_i^T \bfQ_i\|^2.
\end{equation*}
%\begin{equation}
%\phi(\bfH) =\sum_{i=1}^d \frac{1}{2}\|\bfh_i-\bfh_i\bfQ_i\|_F^2
%\end{equation}
Hereabove, $\|\cdot\|_{\rm F}$ denotes the Frobenius norm, and for every $i\in \{1,\ldots,d\}$,
$\lambda_i$ is a positive parameter and
$\bfQ_i$ is a BCCB matrix constructed from the MRF coefficients associated with
the $i$th row of $\bfH$. 
Thus, $\bfQ_i$ enforces possible different spatial structures to $\bfh_1,\cdots,\bfh_d$. 
Note that, because of its form, $\bfQ_i$ can be diagonalized in the frequency domain, i.e., $\bfQ_i=\bfF\bfD_i\bfF^H$,
where $\bfF$ is the 2D FFT matrix and $\bfF^H$ is its inverse.

In the following, we will be interested in the following more general optimization problem:
\begin{equation}
\minimize{\bfH\in \mathbb{R}^{d\times n}} f(\bfH)+ \frac{\gamma}{2} \| \bfH - \overline{\bfH} \|^2_{\rm F}
\label{e:prox}
\end{equation}
where $\gamma \ge 0$ and the second term means that $\bfH$ is close to
$\overline{\bfH}$. When $\gamma = 0$, this problem %\in \mathbb{R}^{d\times n}
reduces to solving \eqref{eq:prob_LS} and, when $\gamma > 0$, this problem 
corresponds to the computation of $\prox_{\gamma^{-1} f}$,
the proximity operator of $\gamma^{-1} f$ \cite{Bauschke2011}. As we will see in the next section, 
such a proximity operator constitutes a key tool for solving optimization problems more involved than \eqref{eq:prob_LS}.
Since $f$ is a quadratic function, it is well-know that $\prox_{\gamma^{-1} f}$ is a linear operator for which a closed-form expression
can be obtained \cite{Combettes2011}. We show next that, rather than applying the direct formula (see \cite[Table 10.1xi]{Combettes2011}), a more efficient approach can be adopted to compute this proximity operator.

Forcing the derivative of the objective function in \eqref{e:prox}
w.r.t. each $\bfh_j$ to be zero and substituting $\bfQ_j=\bfF\bfD_j\bfF^H$ in the resulting equation
leads to 
\begin{equation}
\bfw_j^T\left(\bfW \bfH-\bfY\right)+\lambda_j \bfh_j^T \bfF (\bfI-\bfD_j)^2 \bfF^H + \gamma (\bfh_j - \overline{\bfh}_j)^T = \bf0
\label{eq:deriv_wk_2}
\end{equation}
for every $j\in \{1, \ldots, d\}$. 
Note that the matrix $\lambda_j (\bfI-\bfD_j)^2$ is a real diagonal matrix whose vector of diagonal elements
is denoted by $\bfm_j$.  Thus, \eqref{eq:deriv_wk_2} can be rewritten as
\begin{equation}
\begin{split}
&\bfw_j^T\left(\bfW \bfH-\bfY\right) \bfF + (\bfh_j^T \bfF) \odot \bfm_j^T+ \gamma (\bfh_j - \overline{\bfh}_j)^T \bfF =\bf0
\end{split}
\end{equation}
where $\odot$ is the Hadamard (element-wise) product.
Stacking these $d$ equations leads to the following matrix equation
%\begin{equation}
%\begin{split}
%\bfW^T\left(\bfW \bfH-\bfY\right) \bfF +\lambda (\bfH \bfF) \odot \bfM=0  \\
%\bfS \bfV^T \bfH \bfF + \lambda \bfV^T \left[  (\bfH \bfF) \odot \bfM \right] = \bfV^T \bfW^T \bfY \bfF \\
%(\bfV^T \bfH \bfF) \odot (\bfs \bs{1}^T) + \lambda \bfV^T \left[  (\bfH \bfF) \odot \bfM \right] = \bfV^T \bfW^T \bfY \bfF\\
%\bfV^T \left[ \bfH \bfF \odot (\bfs \bs{1}^T) \right] +\lambda \bfV^T \left[  (\bfH \bfF) \odot \bfM \right] = \bfV^T \bfW^T \bfY \bfF\\
%\bfV^T \left[ \bfH \bfF \odot (\bfs \bs{1}^T+ \lambda \bfM) \right]  = \bfV^T \bfW^T \bfY \bfF\\ % not correct
%\bfH = \left[ \left( \bfW^T \bfY \bfF \right) \oslash (\bs{1} \bfs^T + \lambda \bfM) \right] \bfF^H
%\end{split}
%\end{equation}
%\begin{lemma}
%$\bfS \bfX = \bfX \odot (\bfs \bs{1}^T)$
%\end{lemma}
%
%\begin{lemma}
%$ (\bfA \bfB) \odot ( \bfs \bs{1}^T) = ( \bfA  \odot (\bfs \bs{1}^T) ) \bfB $
%\end{lemma}
%\begin{equation}
%\bfW^T\left(\bfW \bfH-\bfY\right) \bfF +\lambda (\bfH \bfF) \odot \bfM=0
%\end{equation}
%or equivalently 
\begin{equation}
(\bfW^T\bfW+\gamma \bfI) \bfH \bfF  + (\bfH \bfF) \odot \bfM=  (\bfW^T \bfY+\gamma \overline{\bfH}) \bfF.
\label{eq:deri_mat}
\end{equation}
Note that the matrix $\bfM$ can be decomposed as $\bfM=[\bsm_1,\cdots,\bsm_n]$ $=[\bfm_1, \cdots, \bfm_d]^T$,
where a bold italic notation is used to designate the column of $\bfM$ while a bold non-italic one
designates its rows. 
%Similarly, $\bfH=[\bsh_1,\cdots,\bsh_n] =[\bfh_1, \cdots, \bfh_d]^T$.
Eq. \eqref{eq:deri_mat} is a Sylvester-like matrix equation\cite{Wei2015FastFusion, Wei2015CAMSAP, Zhao2016, zhao2016single}\footnote{A Sylvester equation is a matrix equation of the form $\bf AX+XB = C$ \cite{Bartels1972a}.}
w.r.t. $\widetilde{\bfH} =\bf HF$. Let $\widetilde{\bfh}_k$ be the $k$th column of the matrix $\widetilde{\bfH}$ and let $[(\bfW^T \bfY+\gamma \overline{\bfH}) \bfF]_k$ be the $k$th column of $(\bfW^T \bfY+\gamma \overline{\bfH}) \bfF$.
Decomposing \eqref{eq:deri_mat} column-wise allows the estimation of the different vectors $(\widetilde{\bfh}_k)_{1\le k \le n}$
to be decoupled:
%i.e., 
%\begin{equation}
%(\bfW^T\bfW+\gamma \bfI) \widetilde{\bfh}_k+ \lambda \widetilde{\bfh}_k \odot \bsm_k = [(\bfW^T \bfY+\gamma \overline{\bfH}) \bfF]_k,
%\end{equation}
%for every $ k\in \{1, \cdots, n\}$.
%As a consequence, each column $\bar{\bfh}_k$ can be computed analytically as
\begin{equation}
\widetilde{\bfh}_k= \left(\bfW^T\bfW +\gamma \bfI+ \textrm{diag}(\bsm_k)\right)^{-1} [(\bfW^T \bfY+\gamma \overline{\bfH}) \bfF]_k
\label{eq:computetildebfh}
\end{equation}
for every $ k\in \{1, \cdots, n\}$, 
where $\textrm{diag}(\bsm_k)$ is the diagonal matrix whose diagonal is filled with the components of $\bsm_k$.
%\begin{equation}
%\begin{split}
%\bfh_k=& \bfw_k^T\left(\bfY-\sum_{i \neq k} \bfw_i \bfh_i\right)\bfF \\
% &\left[\left(\bfw_k^T\bfw_k\right)\bfI+\lambda_k (\bfI-\bfD_k)^2\right]^{-1}\bfF^H
% \end{split}
%\end{equation}
%\begin{equation}
%\bfh_k= \bfF^H  \bfM_k \bfF \bfr_k
%\end{equation}
%where 
%$\bfM_k=\left[\left(\bfw_k^T\bfw_k\right)\bfI+\lambda (\bfI-\bfD_k)^2\right]^{-1}$
%is a real diagonal matrix and  $\bfr_k = (\bfY-\sum_{i \neq k} \bfw_i \bfh_i^T)^T \bfw_k$ is
%the residual vector. 
The solution to Problem \eqref{e:prox} is finally given by
\begin{equation}
\bfH = \widetilde{\bfH}  \bfF^H.
\end{equation}
If $\max\{d,m\}\ll n$, the computational complexity of the previous strategy
is of the order $\calO(3d n \log_2 n)$ because of the low cost of the 2D-FFT operation. The whole procedure to compute $\prox_{\gamma^{-1} f}(\overline{\bfH})$ is summarized
in Algorithm \ref{Algo:LS_MRF}.

\IncMargin{1em}
\begin{algorithm}[h!]
\label{Algo:LS_MRF}
\KwIn{$\bfY$, $\bfW$, $(\bfQ_i)_{1\le i \le d}$, $\overline{\bfH}$, $\boldsymbol{\lambda}=(\lambda_i)_{1\le i \le d}$, $\gamma$}
%\tcpp{Initializing $\bfH$}
%$\hat{\bfH} \leftarrow \bfH^{(0)}$;\\
\tcpp{2D Fourier diagonalisation of~$(\bfQ_i)_{1\le i \le d}$}
\For{$i=1$ \KwTo $d$}{
$\bfD_i  \leftarrow \bfF^H \bfQ_i \bfF$;  \tcpp{one 2D-FFT required}
%$\bfM_k \leftarrow \left[\left(\bfw_k^T\bfw_k\right)\bfI+\lambda (\bfI-\bfD_k)^2\right]^{-1}$;
$\bfm_i =\lambda_i\textrm{diag}\big((\bfI-\bfD_i)^2\big)$;\\
}
\tcpp{Compute the FFT of $\bfH$ for all pixels in parallel}
\For{$k=1$ \KwTo $n$}{
Compute $\widetilde{\bfh}_k$ using \eqref{eq:computetildebfh};
%$\bar{\bfh}_i \leftarrow \left(\bfW^T\bfW +\lambda \textrm{diag}(\bsm_i)\right)^{-1} (\bfW^T \bfY \bfF)_i$;
}
%\Repeat{convergence}
%{\For{$k=1$ \KwTo $d$}{
%\tcpp{Updating the $k$th row of $\bfH$}
%$ \hat{\bfr}_k \leftarrow (\bfY-\sum_{i \neq k} \bfw_i \hat{\bfh}_i^T)^T \bfw_k$;\\
%$\hat{\bfh}_k  \leftarrow \bfF^H  \bfM_k \bfF \hat{\bfr}_k$;}}
% \bfw_k^T\left(\bfY-\sum\limits_{i \neq k} \bfw_i \hat{\bfh}_i\right)\bfF \left[\left(\bfw_k^T\bfw_k\right)\bfI+\lambda_k (\bfI-\bfD_k)^2\right]^{-1}\bfF^H
$\hat{\bfH} \leftarrow \bar{\bfH}\bfF^H$
\BlankLine
\KwOut{$\hat{\bfH}$}
\caption{Computation of the proximity operator of the LS criterion with GMRF prior}
\DecMargin{1em}
\end{algorithm}

\section{Penalized LS with a GMRF prior}
\label{se:penLS}
Having a fast way of computing the proximity operator of the LS criterion with GMRF prior yields
 efficient solutions to the following broad class of variational formulations:
\begin{equation}
\minimize{\bfH\in \mathbb{R}^{d\times n}} \frac{1}{2} \|\bfY-\bfW \bfH\|_{\rm F}^2 +\sum_{i=1}^d \frac{\lambda_i}{2}\|\bfh_i ^T-\bfh_i^T\bfQ_i \|^2+g(\bfH)
\label{eq:RegLSGMRF}
\end{equation}
where $g\colon\mathbb{R}^{d\times n}\to ]-\infty,+\infty]$ is an additional regularization term, here assumed to be a convex, lower-semicontinuous and proper
function. For example, if $\bfH$ is known to belong to a nonempty closed convex set $\calC \subset \mathbb{R}^{d\times n}$,
a constrained least squares (CLS) regression is obtained by setting $g$ equal to the indicator function of $\calC$, i.e.
\begin{equation}
(\forall \bfU \in \mathbb{R}^{d\times n})\quad
g(\bfU) = \iota_{\calC}(\bfU)=
\left\{
\begin{array}{ll}
0 & \textrm{if } \bfU \in \calC \\
+\infty & \textrm{otherwise.} \end{array} \right.
\label{eq:iotaC}
\end{equation}
Looking for a solution to \eqref{eq:RegLSGMRF} amounts to finding a minimizer of $f+g$. Provided that the proximity operator of 
$g$ is easy to compute, a wide range of proximal algorithms can be employed \cite{Combettes2011,Komodakis_SPM_2015} having good convergence properties. 
In particular, if $g$ is given by
\eqref{eq:iotaC}, this operator reduces to the projection $\Pi_{\calC}$ onto $\calC$. 

As an example of proximal approaches which can be used,
Algorithm \ref{Algo:Fusion} describes the iterative steps to be followed in order to implement the alternating direction of multipliers method (ADMM) \cite{Bioucas2010SUNSAL,Boyd2011}.

\IncMargin{1em}
\begin{algorithm}[h!]
\label{Algo:Fusion}
\KwIn{$\bfY$, $\bfW$, $(\bfQ_i)_{1\le i \le d}$, $\bfU^{(0)}$, $\bfG^{(0)}$, $\boldsymbol{\lambda}$, $\gamma$}
\tcpp{Initialize $\bfU$ and $\bfG$ with $\bfU^{(0)}$ and~$\bfG^{(0)}$} 
$\hat{\bfU} \leftarrow \bfU^{(0)}$;\\
$\hat{\bfG} \leftarrow \bfG^{(0)}$;\\
\tcpp{ADMM iterations}
\Repeat{convergence}
{
Update $\hat{\bfH}$: $\hat{\bfH}  \leftarrow \prox_{\gamma^{-1} f}(\hat{\bfU}+\hat{\bfG})$\\
by feeding Algorithm \ref{Algo:LS_MRF} with\\
 ($\bfY$, $\bfW$, $(\bfQ_i)_{1\le i \le d}$, $\hat{\bfU}+\hat{\bfG}$, $\bs{\lambda}$, $\gamma$);\\
Update $\hat{\bfU}$: $ \hat{\bfU} \leftarrow \prox_{\gamma^{-1}g}(\hat{\bfH}-\hat{\bfG})$;\\
Update $\hat{\bfG}$: $ \hat{\bfG} \leftarrow \hat{\bfG}-\hat{\bfH}+\hat{\bfU}$;
}
\BlankLine
\KwOut{$\hat{\bfH}$}
\caption{Penalized LS with GMRF regularization}
\DecMargin{1em}
\end{algorithm}

\begin{figure}[htb!]
\centering
\includegraphics[width=0.7\columnwidth]{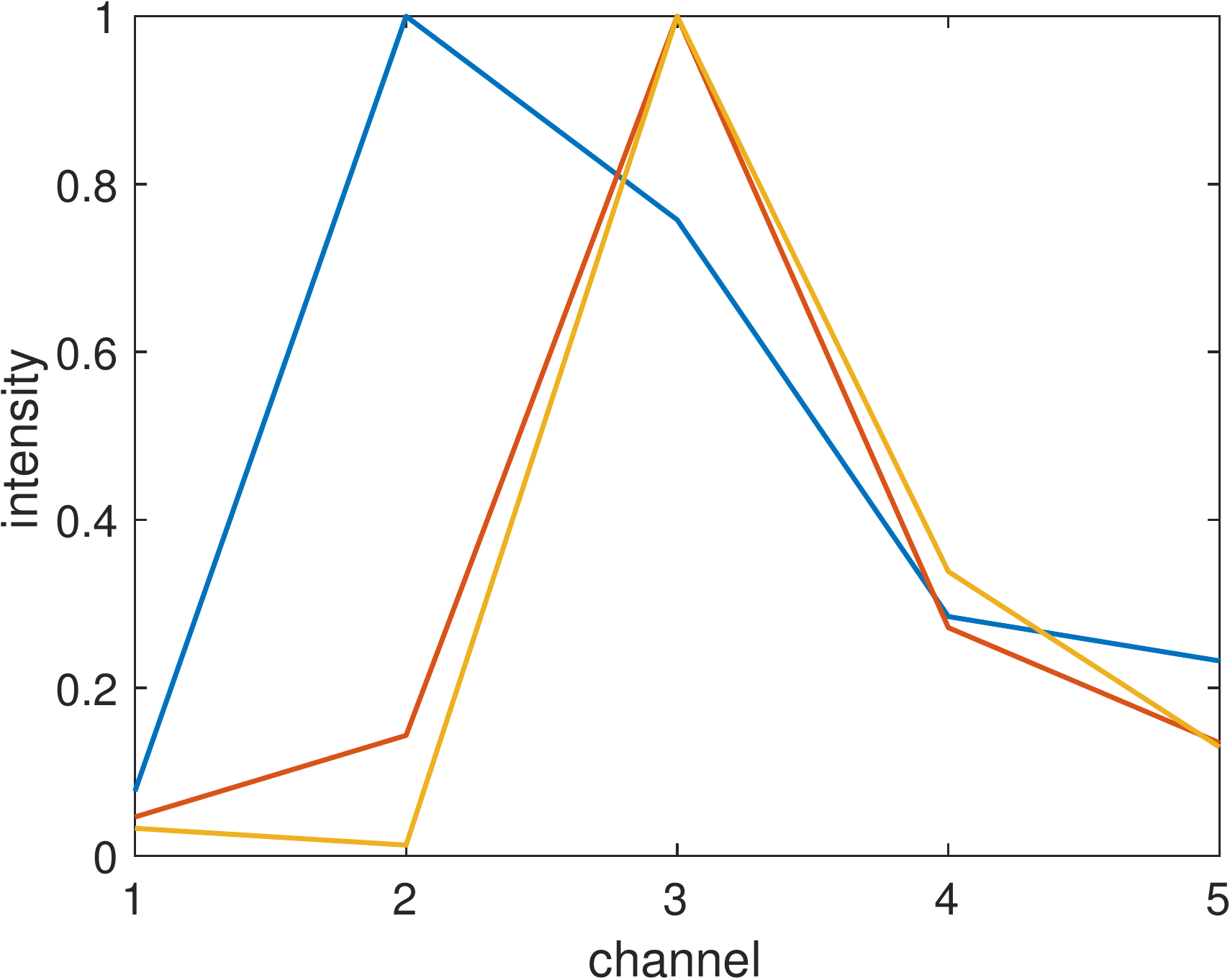}
\caption{Columns of the matrix $\bfW$.}
\label{fig:basis}
\end{figure}

\section{Experiments}
\label{sec:simu}
This section evaluates the performance of our algorithm for a multichannel image processing problem, 
and compares it with two widely used optimization algorithms: forward backward (FB) \cite{Combettes2005}
and FISTA \cite{Beck2009}. For a fair comparison, all the algorithms have been implemented using 
MATLAB R2016b on an HP EliteBook Folio 9470m with Intel(R) Core(TM) i7-3687U CPU @2.10GHz and 16GB RAM.
% a MacPro computer with Intel(R) Core(TM) i7 CPU@2.70GHz and 16GB RAM.
%HP EliteBook Folio 9470m, Intel(R) Core(TM) i7-3687U CPU @2.10 GHz, 16GB RAM

\subsection{Simulation scenario}
In all the experiments, we %have three bases (columns) and five channels (rows) 
consider a matrix $\bfW \in \mathbb{R}^{5 \times 3}$ corresponding to measurements
acquired in five channels and decomposed in a basis defined by three vectors.
The three columns of the basis matrix $\bfW$ are displayed in Fig. \ref{fig:basis}. 
These vectors represent the signatures\footnote{courtesy of Alexandre Jaouen, CNRS-AMU UMR7289.} 
of three different fluorescent protein spectra \cite{Ricard2014}. One can note that two of them (red and brown) are quite similar, which makes the 
model very ill-posed.
%The matrix $\bfH$ has been generated according to two scenarios.
The matrix $\bfH$ has been generated row by row after vectorizing $3$ texture images available at \url{http://sipi.usc.edu/database/}.
The three images we have considered in this work are displayed in the first row of Fig. \ref{fig:real_H}
showing clear oriented structures. The GMRF parameters for these three images have been estimated 
using the maximum likelihood method \cite{Won1988} and are summarized in Fig. \ref{fig:GMRF_coeff_real}. 
Note that these GMRFs consider $3 \times 3$ neighbors  around 
one pixel and that half of them are set to zeros due to the symmetry property.
The size of the images is $512 \times 512$.
In our simulations, the regularization parameters $(\lambda_i)_{1\le i \le 3}$ for all bands are 
chosen equal to $0.05$ empirically (in real application this value vary depending on the noise power).
The convex penalty function $g$ is the indicator of the box constraint $\bfH \in [0,1]^{d\times n}$.
%of a box constraint such as $0 \leq \bfH \leq 1$,
%where $\leq$ is understood in element-wise sense.

%\subsubsection{Random regression coefficients}
%The rows of the matrix $\bfH$ have been generated using three
%GMRF distributions characterized by the regression coefficients
%indicated in Fig. \ref{fig:GMRF_coeff_random}.
%Note that these GMRFs consider $3 \times 3$ neighbors  around 
%one pixel and that half of them are set to zeros due to the symmetric property.
%The resulting ground-truth images of size $190 \times 190$ are shown in
%Fig. \ref{fig:random_H}. The size of the generated image is $190 \times 190$.
% 
%\begin{figure}
%\[
%\begin{bmatrix}
%0.5 & 0.4 & 0\\
%-0.2 & 0 & 0\\
%0 & 0 & 0\\
%\end{bmatrix}
%\begin{bmatrix}
%0.1 & 0 & 0\\
%0.8 & 0 & 0\\
%-0.08 & 0 & 0\\
%\end{bmatrix}
%\begin{bmatrix}
%0.5 & 0 & 0\\
%0.5 & 0 & 0\\
%-0.25 & 0 & 0\\
%\end{bmatrix}
%\]
%\caption{Generated GMRF coefficients for $\bfh_1$, $\bfh_2$ and $\bfh_3$ (left to right).}
%\label{fig:GMRF_coeff_random}
% \end{figure}

%\subsubsection{Real images}
%Another way to generate $\bfH$ is to vectorize real images to construct the rows of $\bfH$.

The observed data are finally generated using the linear mixing model \eqref{eq:linear_model},
i.e., $\bf Y=WH+N$, where the noise matrix $\bfN$ has been generated 
using samples of a Gaussian distribution with zero mean and covariance matrix $\sigma^2 \bfI$.
The variance $\sigma^2$ has been adjusted in order to have an initial SNR
(signal to noise ratio)  equal to $25$dB.

\begin{figure}[htb!]
\[
\begin{bmatrix}
-0.26 & 0.55 & 0\\
0.13 & 0 & 0\\
 0.58 & 0 & 0\\
\end{bmatrix}
\begin{bmatrix}
 -0.19 & 0.78 & 0\\
 0.35 & 0 & 0\\
 0.042 & 0 & 0\\
\end{bmatrix}
\begin{bmatrix}
-0.68 & 0.79  & 0\\
0.84 & 0 & 0\\
0.047 & 0 & 0\\
\end{bmatrix}
\]
\caption{Estimated GMRF coefficients for $\bfh_1$, $\bfh_2$ and $\bfh_3$ (left to right).}
\label{fig:GMRF_coeff_real}
\end{figure}

\vspace{-0.3cm}
\subsection{Quality Assessment}
To analyze the quality of the proposed estimation method, we have considered the
normalized mean square error (NMSE) defined as %\cite{Bioucas2010SUNSAL}
\begin{align*}
    \textrm{NMSE} = \frac{\|\hat{\bfH}- \bfH\|^2_{\rm F}}{\| \bfH\|^2_{\rm F}}.
\end{align*}
The smaller \textrm{NMSE}, the better the estimation quality.

\subsection{Comparison with existing optimization algorithms}
\label{subsec:simulation}
\begin{figure}
\centering
\includegraphics[width=0.24\textwidth]{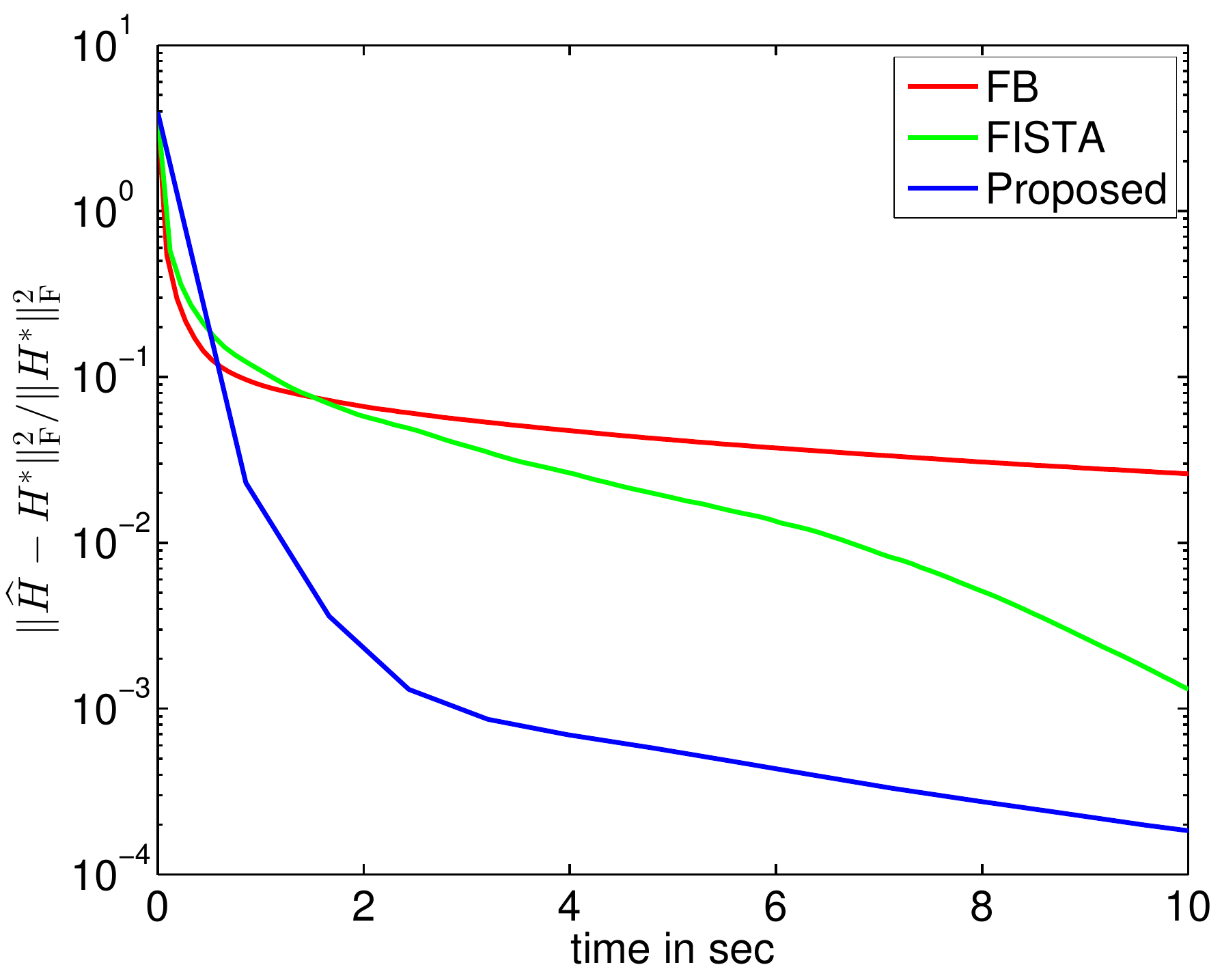}
\includegraphics[width=0.24\textwidth]{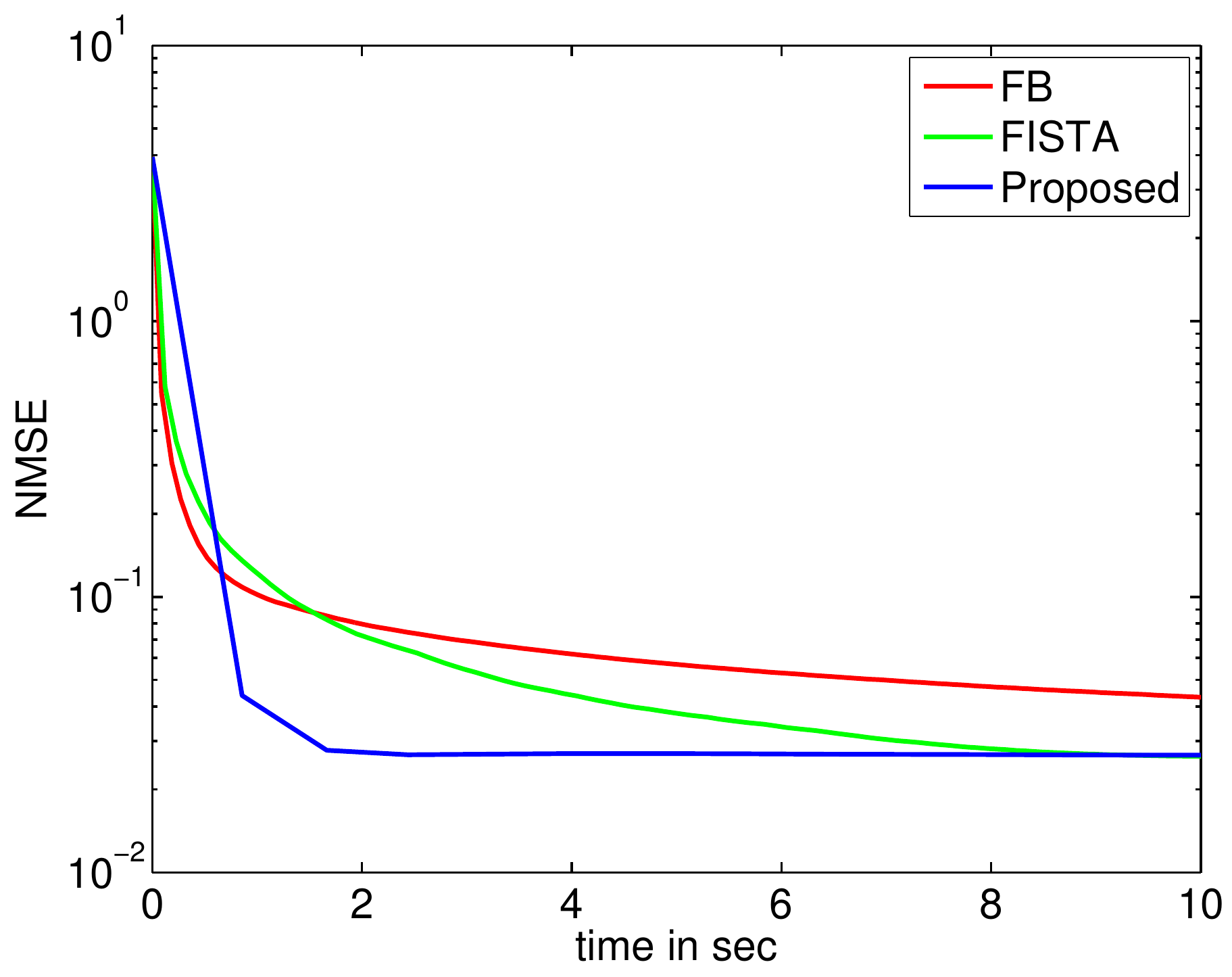}
\caption{Convergence comparison of different algorithms: (left) relative distance to the solution \emph{vs} time,
(right) NMSE \emph{vs} time.}
\label{fig:comparison}
\end{figure}

The evolution of the relative error between the iterates and the solution to \eqref{eq:RegLSGMRF} versus
execution time, is displayed in Fig. \ref{fig:comparison}(left) for the three tested algorithms, namely FB, FISTA 
and the proposed one. Here, the optimal solution $\mathbf{H}^*$ has been precomputed for each algorithm 
using a large number of iterations. We also show the NMSE versus time in Fig. \ref{fig:comparison}(right). 
All the algorithms lead to the same estimation quality as expected. However, 
as demonstrated in these plots, the proposed algorithm based on 
a Sylvester-like equation solver is faster than FB and FISTA. 
More precisely, the proposed algorithm converges rapidly in a few steps while the other two need more iterations and time
to converge. One can also note that FISTA converges faster than FB, both in terms of error on the iterates and NMSE decays.

%\begin{figure*}
%\centering
%\includegraphics[width=0.3\textwidth]{figures_MRF/cost_time_zoom}
%\includegraphics[width=0.3\textwidth]{figures_MRF/cost_iters}\\
%\includegraphics[width=0.3\textwidth]{figures_MRF/error_time}
%\includegraphics[width=0.3\textwidth]{figures_MRF/error_iters}
%\caption{Convergence comparison of different algorithms: (top) objective function \emph{vs} (time $\&$ iterations),
%(bottom) NMSE \emph{vs} (time $\&$ iterations).}
%\label{fig:comparison}
%\end{figure*}

\begin{figure}
\centering
\begin{tabular}{@{}ccc@{}}
\includegraphics[width=0.3\columnwidth]{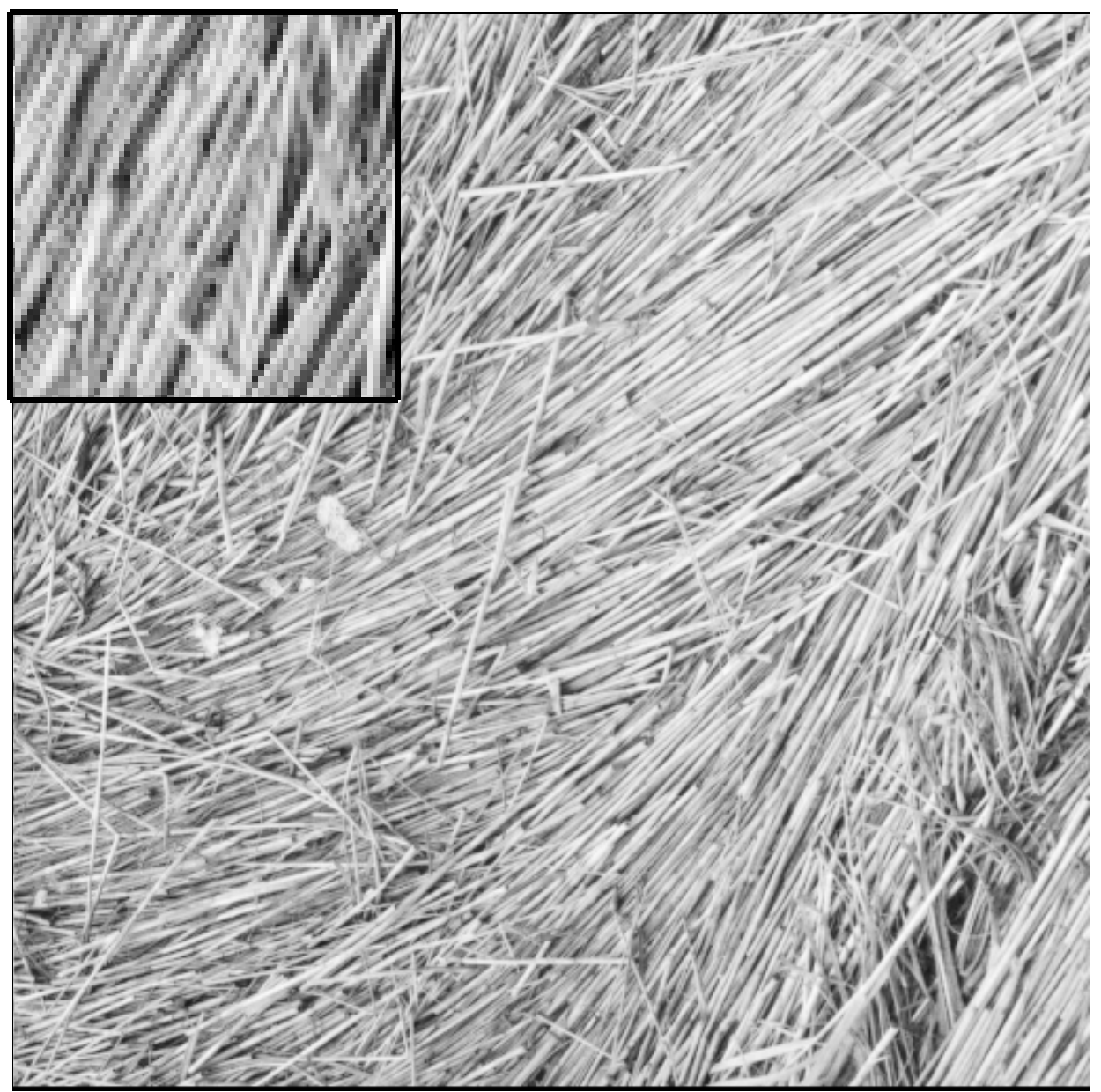}&
\includegraphics[width=0.3\columnwidth]{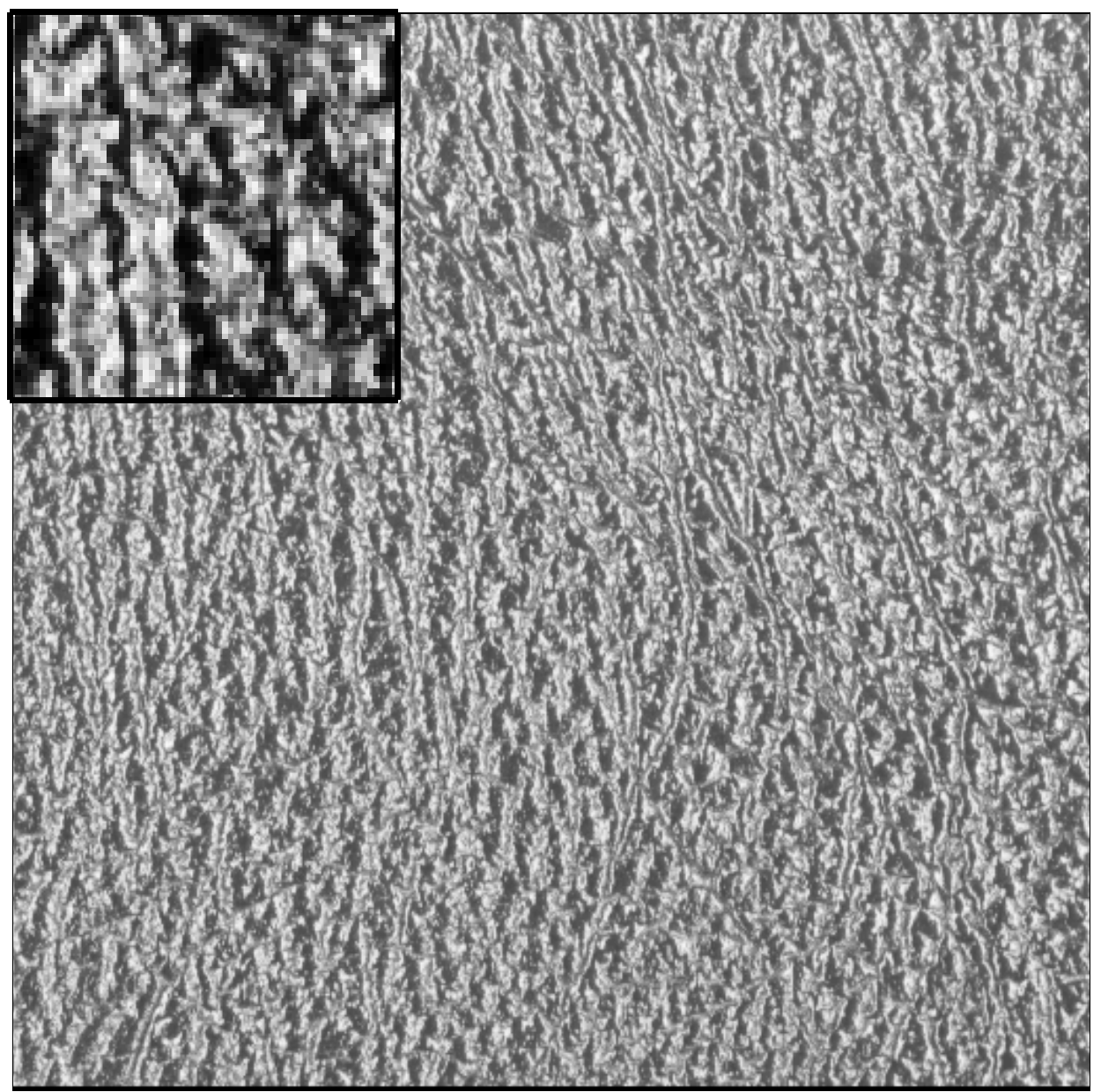}&
\includegraphics[width=0.3\columnwidth]{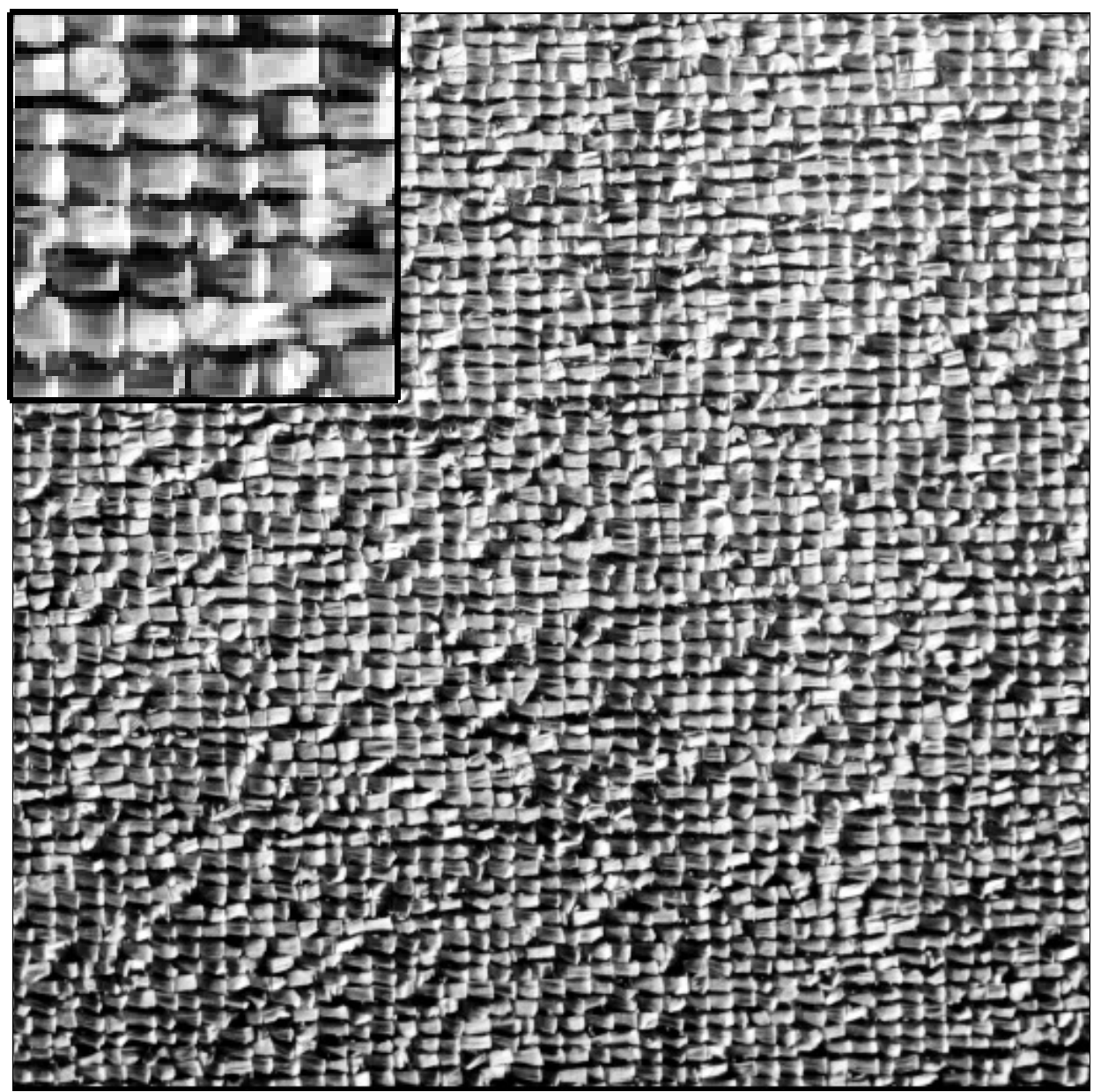}\\
\includegraphics[width=0.3\columnwidth]{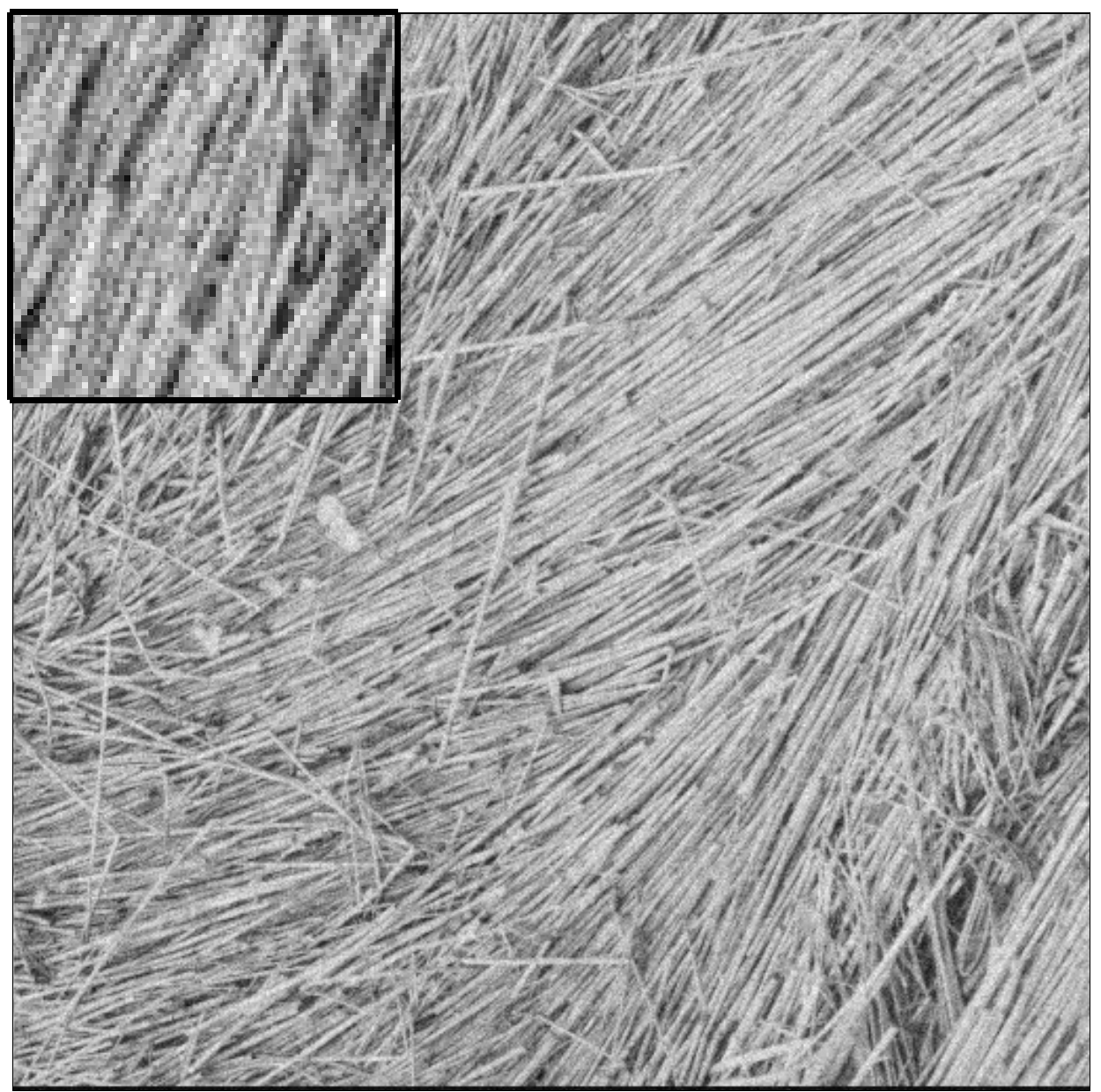}&
\includegraphics[width=0.3\columnwidth]{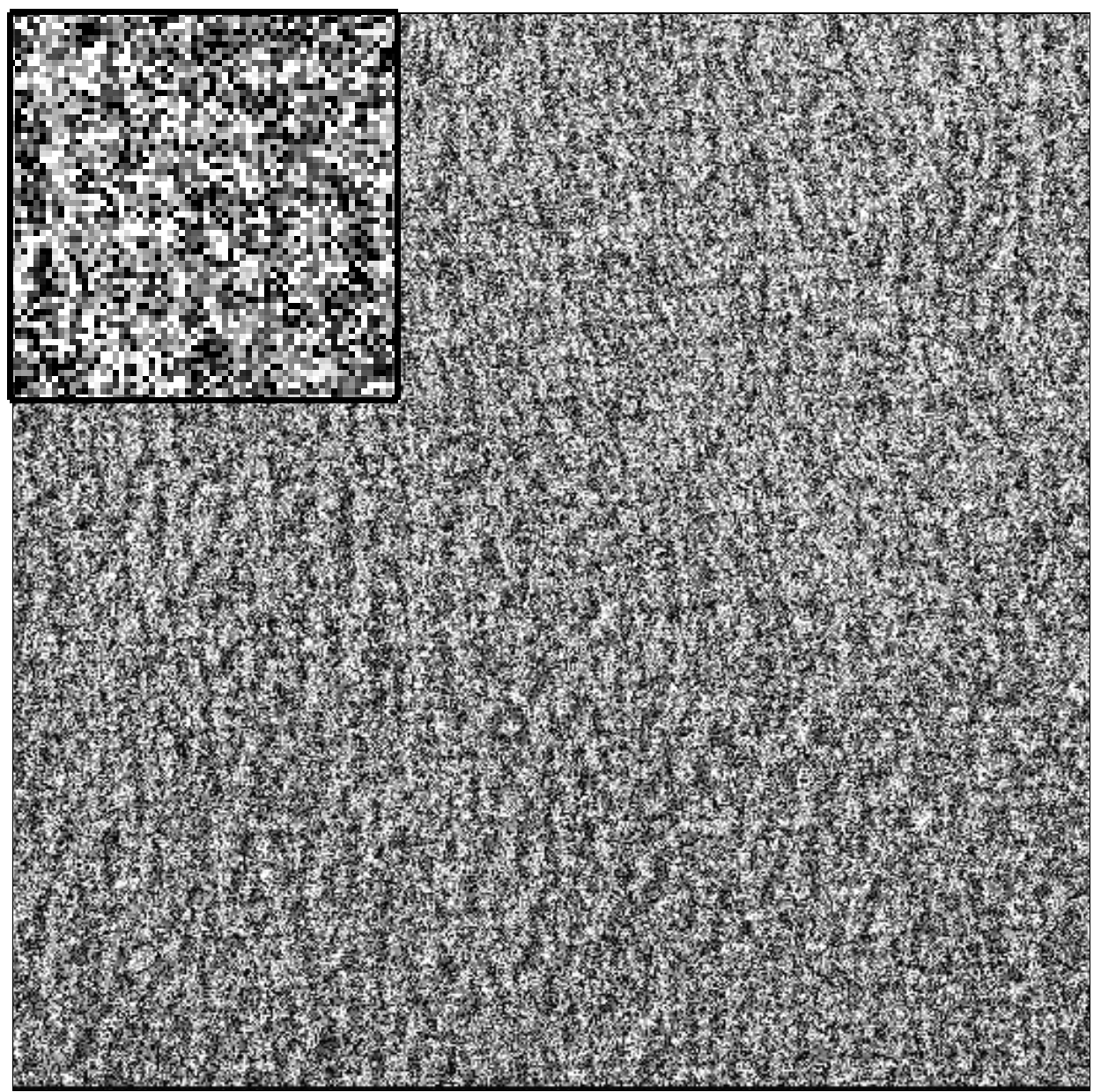}&
\includegraphics[width=0.3\columnwidth]{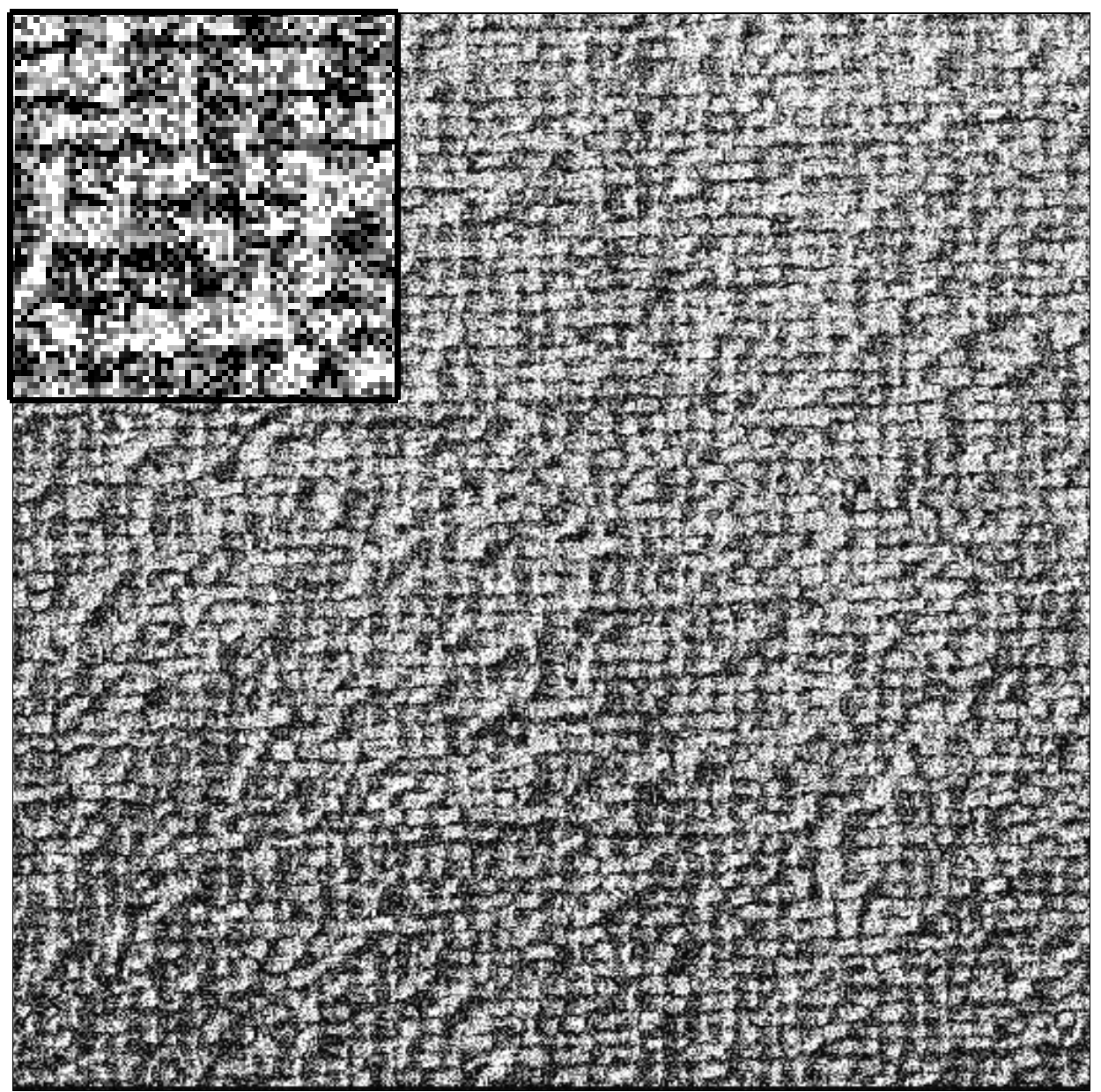}\\
\includegraphics[width=0.3\columnwidth]{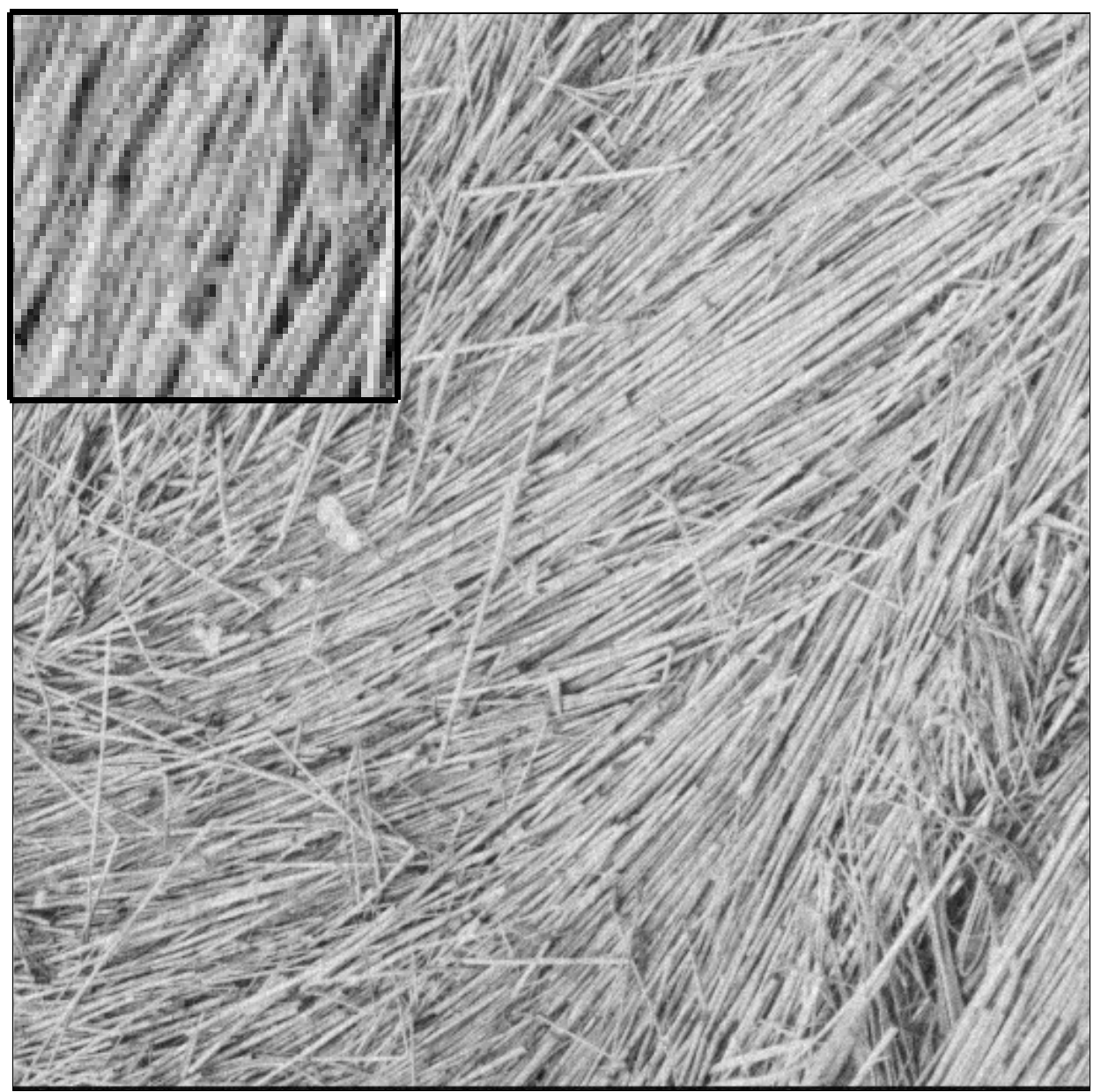}&
\includegraphics[width=0.3\columnwidth]{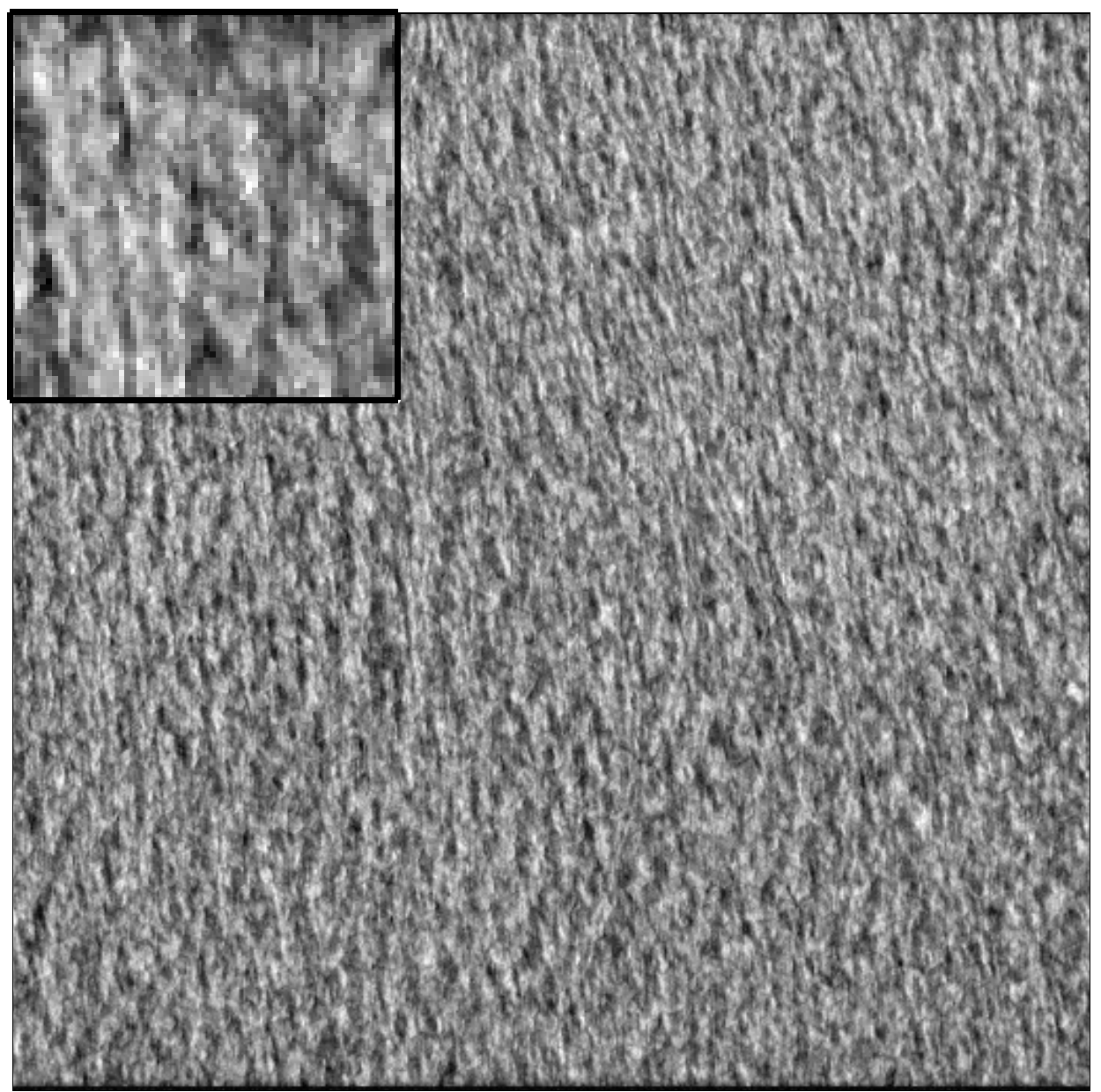}&
\includegraphics[width=0.3\columnwidth]{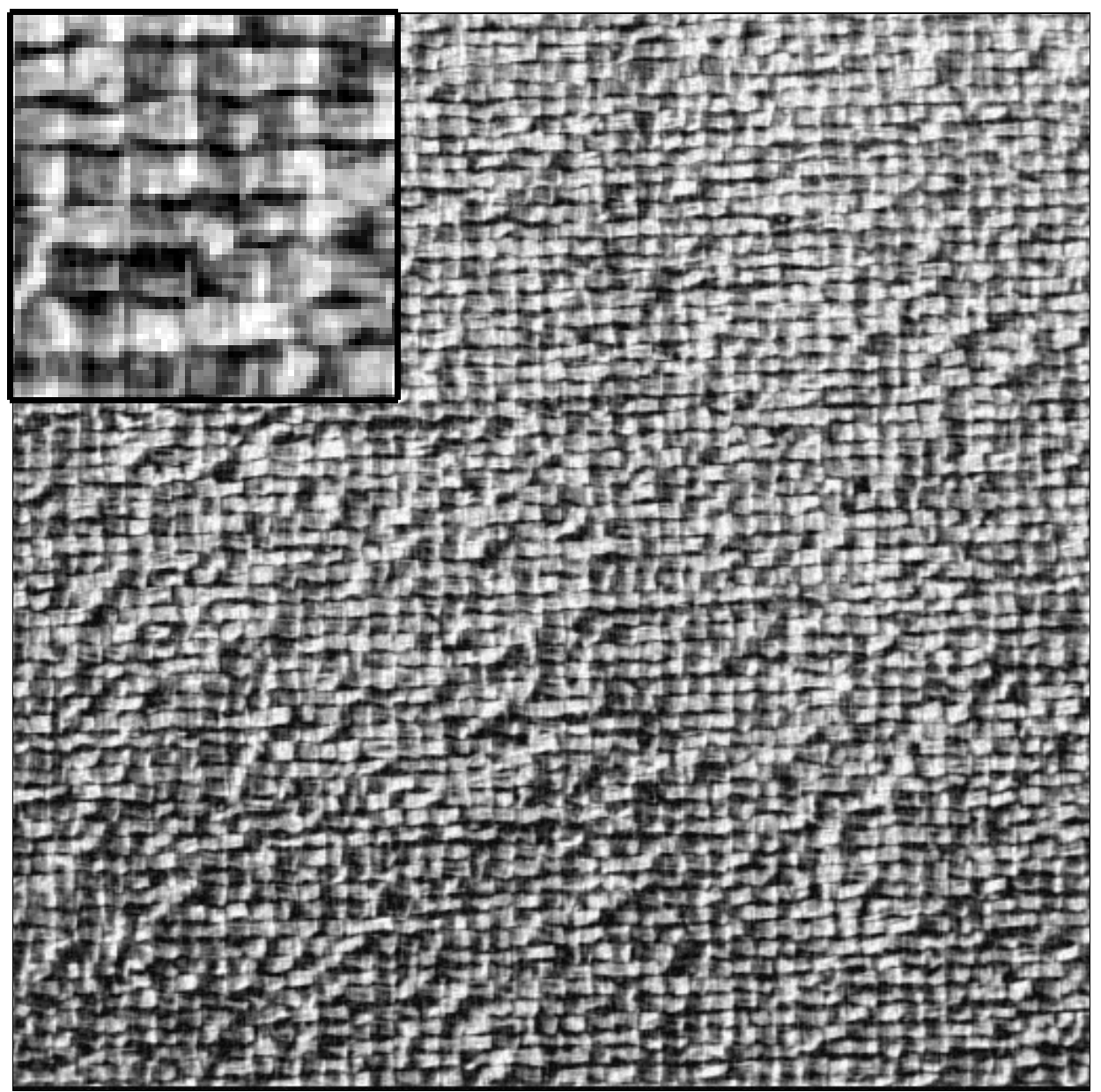}
\end{tabular}
\caption{Regression matrix $\bfH$ and its estimates $\hat{\bfH}$ for texture images ($512 \times 512$): (top) ground-truth, (middle) LS estimator (NMSE = $0.166$), (bottom) LS estimator with GMRF (NMSE = $0.026$). A zoom of each image is displayed in its left top side.}
\label{fig:real_H}
\end{figure}

To demonstrate the role of the GMRF regularization, we computed the box constrained ($\bfH \in [0,1]^{d\times n}$)
LS regression without any regularization,
by setting $\lambda_i=0$ for every $i\in\{1,2,3\}$ and use it as a baseline for comparison. 
The regression matrix $\bfH$ estimated by LS and by the proposed approach
are displayed in the second and third rows of Fig. \ref{fig:real_H}, respectively.
Due to the ill-posedness of the problem, the inversion without any spatial regularization 
amplifies the noise, leading to poor estimation results as shown in the second row 
of Fig. \ref{fig:real_H} (especially for the second and third images).
The GMRF model plays a very important role in
restoring satisfactorily the spatial structures and details as shown in the last row
of Fig. \ref{fig:real_H}. The NMSE values indicated in the caption of Fig. \ref{fig:real_H}
corroborate these visual comparisons.

\vspace{-0.2cm}
% Obviously,
%the results given by the proposed algorithm are much better than the ones without
%any regularization, both qualitatively and quantitatively. 
%The performance w.r.t. $\lambda$ is given in Fig. \ref{fig:real_lambda},
%which shows that the highest RSNR is obtained when $\lambda=0.04$.

%\subsection{Selection of the regularization parameter $\lambda$}
%To select an appropriate value of $\lambda$, the performance of the proposed algorithm has been evaluated 
%as a function of $\lambda$. The results are displayed in Fig. \ref{fig:random_lambda} showing
%that the RSNR reaches the highest value $11.97$dB when $\lambda=0.0025$ 
%as annotated in the figure. It is noteworthy that in a wide range of $\lambda$,
%the proposed method always outperforms the LS estimator without regularization.
%
%\begin{figure}
%\centering
%\includegraphics[width=0.8\columnwidth]{figures_MRF/real/lambda}
%\caption{RSNR of the proposed fusion algorithm versus $\lambda$.}
%\label{fig:real_lambda}
%\end{figure}

%\vspace{-0.1cm}
\section{Conclusion}
\label{sec:conclusion}
This paper developed a new algorithm for penalized least squares regression with GMRF
regularization based on a Sylvester-like matrix equation solver.
The closed-form solution of this equation makes it very appealing in terms of computational 
complexity. Although we have focused on the use of ADMM,
the proposed approach can be embedded into most of the existing proximal methods
to solve penalized or constrained least squares regression problems. Numerical experiments confirmed 
the effectiveness of the resulting algorithms. Future work includes the generalization
of the proposed algorithm to applications where the basis matrix is partially known or unknown.

\section*{Acknowledgment}
The authors thank CNRS for supporting this work 
by the CNRS Imag'In project under grant 2015 OPTIMISME.
%This work was supported by the CNRS Imag’In project under grant 2015 OPTIMISME.

\bibliographystyle{ieeetran}
\bibliography{strings_all_ref,biblio_all}
%%\bibliography{D:/qwei2/Dropbox/Latex/strings_all_ref,D:/qwei2/Dropbox/Latex/biblio_all}
%\bibliography{/Users/qw62/Dropbox/Latex/strings_all_ref,/Users/qw62/Dropbox/Latex/biblio_all}
\end{document}